\documentclass[letterpaper, 10 pt, conference]{IEEEtran}  

\IEEEoverridecommandlockouts                              

\usepackage{graphics} 
\usepackage{epsfig} 
\usepackage{mathptmx} 
\usepackage{times} 
\usepackage{amsmath} 
\usepackage{amssymb}  
\usepackage{hyperref}
\usepackage{float}
\usepackage{multirow}
\usepackage[font=small, skip=4pt]{caption}
\usepackage{cite}
\usepackage{empheq}

\newcommand\ddfrac[2]{\frac{\displaystyle #1}{\displaystyle #2}}

\title{\LARGE \bf
Optimal Combination Forecasts on Retail Multi-Dimensional Sales Data}

\author{\IEEEauthorblockN{Luis Roque}\IEEEauthorblockA{Faculty of Engineering, University of Porto\\HUUB AI Lab\\Porto, Portugal\\laroque@thehuub.co}\and\IEEEauthorblockN{Cristina A. C. Fernandes}\IEEEauthorblockA{HUUB AI Lab\\Porto, Portugal\\cafernandes@thehuub.co}\and\IEEEauthorblockN{Tony Silva}\IEEEauthorblockA{HUUB AI Lab\\Porto, Portugal\\cafernandes@thehuub.co}}

\begin{document}

\maketitle
\thispagestyle{plain}
\pagestyle{plain}

\begin{abstract}

Time series data in the retail world are particularly rich in terms of dimensionality, and these dimensions can be aggregated in groups or hierarchies. Valuable information is nested in these complex structures, which helps to predict the aggregated time series data. From a portfolio of brands under HUUB's monitoring, we selected two to explore their sales behaviour, leveraging the grouping properties of their product structure. Using statistical models, namely SARIMA, to forecast each level of the hierarchy, an optimal combination approach was used to generate more consistent forecasts in the higher levels. Our results show that the proposed methods can indeed capture nested information in the more granular series, helping to improve the forecast accuracy of the aggregated series. The Weighted Least Squares (WLS) method surpasses all other methods proposed in the study, including the Minimum Trace (MinT) reconciliation.

\end{abstract}

\section{INTRODUCTION}

There are numerous potential benefits of using forecasting models on fashion companies, such as a reduction of the bullwhip effect, a possible reduction of the difficulties of the supplier production, an effectiveness improvement of the sourcing strategy of the retailer, the reduction of lost sales and markdowns and, consequently, the increase of profit margins, as described by \cite{Lenort2013}.

Fashion markets are, however, volatile and hard to predict. \cite{Lenort2013} points out the different perspectives and requirements to bear in mind when developing a forecasting model to suit its specificities. There are two main forecast horizons to consider: medium term to plan sourcing and production and short term to  plan replenishment. The products life cycle is very short in the fashion industry and there are some additional constraints such as "one shot" supply \cite{Lenort2013}. This shows the relevance of forecasting in the purchase quantities determination. At the same time, there are basic and best selling items with a completely different life and purchasing cycles. Due to these short life cycles, historical data of the disaggregated product structure are ephemeral, hence the importance of forecasting different levels of aggregation. The strong seasonality also arises as one of the important behaviours to model. The sales data generated by companies is therefore highly dimensional, meaning that it is possible to aggregate and disaggregate it by a significant number of dimensions, e.g product structure, geography, etc. 

Several studies show that forecasting the aggregates by disaggregates results in better forecasts than using the best individual models \cite{Capistran2010,athanasopoulos2009hierarchical,Shang2017}. The most common approach to leverage the information present in each aggregation level, hierarchy or group, was initially to use the top-down or the bottom-up approaches. \cite{hyndman2011optimal} proposed a new method for combining different levels coherently across the aggregation structure, resulting in what they designated \textit{optimal forecasts}, which proved better than the standard methods. The work by \cite{athanasopoulos2009hierarchical} was one of the first applications of the hierarchical forecasting model and its focus was on tourism demand time series data. The data were disaggregated by geographical region and by purpose of travel, thus forming a natural hierarchy of time series to reconcile. The potential of this optimal combination method was demonstrated immediately in the first attempt, surpassing the bottom-up and all the top-down approaches, with the exception of the top-down approach based on forecast proportions. Later, \cite{hyndman2016fast} presented improvements in the method and proved that the optimal combination method actually outperformed all the traditional methods. 

Another important discussion in the field is the comparison between statistical models and machine learning. \cite{Makridakis2018} shared the findings on the M3 competition, where a large subset of 1045 monthly time series was used, and statistical methods dominated across all the accuracy measures and forecasting horizons. More recently, the M4 competition has shown the potential of hybrid and combination models, using both statistical and machine learning models, which dominated the top 10 of the competition \cite{Makridakisetal2018}. 

Our work is focused on statistical models to build solid baseline forecasting models and optimally reconcile them using a hierarchical approach, but also sustained by the work being done in the field, where machine learning models still do not match the performance of statistical models when only a pure time series is provided. 

Section~\ref{sec:related_work} describes the theory and previous studies related to the methodologies implemented in this work. Section~\ref{sec:performance_measures} lists the performance measures used to analyse and compare our models. In Section~\ref{sec:dataset} we present the dataset in study and in Section~\ref{sec:methods} we describe the methodology applied. Our results and discussion are detailed in Section~\ref{sec:Results} and our conclusions and future work are summarised in Section~\ref{sec:conclusion}.


\section{RELATED WORK}\label{sec:related_work}

\subsection{Data Pre-processing}

Different approaches can be taken to pre-process time series data in order to stationarize, normalize, detrend and deseasonalize data. Pre-processing is an important stage as it has been shown that forecasting models perform better on pre-processed data \cite{Makridakisetal2018}. Indeed, a time series will be easier to model accurately the more stationary and closer to a normal distribution it is. For a time series to be stationary, its properties, such as mean, variance and autocorrelation, should be constant over time. Various transformations can be applied to data for this purpose.
The Box-Cox transformation is used to modify the distributional shape of a set of data to stabilize variance, make the data more normal distribution-like so that tests and confidence limits that require normality can be appropriately used. It is defined by:
    \begin{equation}
        y^{(\lambda)} = 
        \begin{cases}
    (y^{\lambda}-1)/\lambda &, \text{if } \lambda \neq 0  \\
    \log(y) &, \text{if } \lambda=0   
    \end{cases}            
    \end{equation}
where $\lambda$ is the transformation power parameter and $y$ is a list of strictly positive numbers. Hence, the Box-Cox transformation can only be applied to positive data.

A Box-Cox transformation with $\lambda=0$ is equivalent to a simple logarithmic transformation.


\subsection{Model description}

The Autoregressive Integrated Moving Average (ARIMA) or Box Jenkins methodology is one of the most popular and frequently used stochastic time series models. The time series is assumed to be linear and follows a particular known statistical distribution, such as the normal distribution, which can make it inadequate in some practical situations \cite{2013arXiv1302}. However, its flexibility to represent several varieties of time series with simplicity and the optimization of the model building process render the Box-Jenkins methodology as a well-suited baseline approach \cite{2013arXiv1302}.

In an AR$(p)$ model, the future value of a variable is assumed to be a linear combination of $p$ past observations and a random error together with a constant term \cite{2013arXiv1302}. Mathematically:
\begin{equation}
    y_{t}  =  c + \sum_{i=1}^{p}{\phi_{i} y_{t-i}} + \epsilon_{t},
\end{equation}
where $y_{t}$ is the model value; $\epsilon_{t}$ is the random error (or random shock) at time period $t$; $\phi_{i}$ are model coefficients and $c$ is a constant; $p$ is the order of the model, also called the \textit{lag}. 





\medskip
The MA$(q)$ model uses past errors as the explanatory variables \cite{2013arXiv1302}. Mathematically:
\begin{equation}
    y_{t}  =  \mu + \sum_{j=1}^{q} \theta_{j} \epsilon_{t-j} + \epsilon_{t},
\end{equation}
where $\mu$ is the mean of the series; $\theta_{j}$ are the model parameters; $q$ is the order of the model. The random errors, or random shocks, are assumed to be white noise, i.e. independent and identically distributed (i.i.d.) random variables with zero mean and a constant variance $\sigma^{2}$ \cite{2013arXiv1302}. 


\medskip
ARMA is a combination of the autoregressive and moving average models. Matematically:
\begin{equation}
    y_{t}  =  c + \epsilon_{t} + \sum_{i=i}^{p}{\phi_{i}y_{t-i} + \sum_{j=1}^{q}\theta_{j} \epsilon_{t-j}}
\end{equation}

The ARMA model can only be used for stationary time series data. However, in practice many time series such as those related to socio-economy and business show non-stationary behaviour. Time series, which contain trend and seasonal patterns, are also non-stationary in nature. For this reason, the ARIMA model is proposed.

The ARIMA model is a generalization of an ARMA model to include the case of non-stationarity. The non-stationary time series are made stationary by applying finite differencing of the data points. Mathematically it can be written as:
\begin{equation}
    \begin{aligned}
        & (1-\sum_{i=1}^{p}{\phi_{i}L^{i}})(1-L)^{d}y_{t} = (1+\sum_{j=1}^{q}{\theta_{j}L^{j}})\epsilon_{t} \\
        & \Leftrightarrow \phi(L)(1-L)^{d}y_{t} = \theta(L) \epsilon_{t} \\
    \end{aligned}
\end{equation}
where $L$ is the lag or backshift operator, defined as $Ly_{t} = y_{t-1}$ and 
\begin{equation}
     \begin{aligned}
         \phi(L) &= 1-\sum_{i=1}^{p}{\phi_{i}L^{i}}\\
         \theta(L) &= 1 + \sum_{j=1}^{q}{\theta_{j}L_{j}},
    \end{aligned}
\end{equation}
with $p$, $d$ and $q$ being integers that refer to the order of the autoregressive, integrated and moving average parts of the model, respectively. The integer $d$ controls the level of differencing. 


\medskip
The Seasonal Autoregressive Integrated Moving Average (SARIMA) model is the generalization of the ARIMA model for seasonality. In this model, the seasonality is removed by seasonally differencing the series. In other words, by computing the difference between one observation and the corresponding observation from the previous year (or season): $z = y_{t} - y_{t-s}$. For a monthly time series, $s=12$, for a quarterly time series $s=4$, etc.

Mathematically, a SARIMA $(p,d,q) \times (P,D,Q)^{s}$ model is formulated as:
\begin{equation}\label{eq:sarima}
    \Phi_{P}(L^{s}) \; \phi_{p}(L) \; (1-L)^{d} \; (1-L)^{D} \; y_{t} = \Theta_{Q}(L^{s}) \; \theta_{q}(L) \; \epsilon_{t} \\
\end{equation}
where $P$, $D$, $Q$ are the seasonal orders of the autoregressive, integrated and moving average parts of the model, respectively, and $s$ is the seasonal period.

\subsection{Model Evaluation}

Good models are obtained by minimising the Akaike’s Information Criterion (AIC), the Corrected Akaike’s Information Criterion (AIC$_{c}$) or the Schwarz’s Bayesian Information Criterion (BIC). The BIC and the AIC$_{c}$ were developed as a bias-corrected version of the AIC, better fit for short time series, where the AIC tends to select more complex models with too many predictors. \cite{hyndman2018forecasting} indicates its preference to use the AIC$_{c}$ to compare models.

A note should be made about the fact that the AIC, AIC$_{c}$ and BIC criteria should not be used to compare the performance of models with different differencing orders. The reason for this is that these criteria are all based on the likelihood of the model. This likelihood will be different for a series that is differenced and for a series that is not differenced. These criteria can only directly compare models with the same seasonal and first differencing orders. 

\medskip

If there are significant correlations between different lags of the data, or the residuals of a model, this indicates that there are still important features in the data that the model is not reproducing well and thus there is valid information to learn. \cite{ljungbox1978} proposed the Ljung-Box (LB) test to evaluate the overall randomness of data based on a number of lags. It is commonly used to check if the residuals from a time series resemble white noise, i.e. if the relevant information was captured by the model. It is based on the statistic $Q^{*}$, which can be written as:
\begin{equation}
    Q^{*} = T(T + 2) \sum_{k}^{l}(T - k)^{-1} \, r^2_k  , 
\end{equation}
where $T$ is the length of the time series, $r_{k}$ is the $k^{th}$ autocorrelation coefficient of the residuals and $l$ is the number of lags to test. $Q^{*}$ approximates a chi-squared distribution with $l - K$ degrees of freedom, where $K$ is the number of parameters of the estimated model. Large values of $Q^{*}$ indicate that there are significant autocorrelations in the residual series. 

The null hypothesis of the LB test states that the data are independently distributed, not showing serial correlation. A significant p-value in this test thus rejects the null hypothesis. Conversely, small values of p-value indicate the possibility of non-zero autocorrelations.

\subsection{Hierarchical and Grouped time series}

Univariate Time Series data rely only on time to express patterns. When there is aggregated data available, there are potential covariance patterns nested in the hierarchy. \cite{fliedner2001hierarchical} summarizes guidelines on using the more traditional hierarchical forecasting approaches, top-down, bottom-up and a combination of both, the middle-out approach. Among these three, top-down and middle-out rely on a unique hierarchy to assign the weights from the higher aggregated series to the lower ones, so it is not suitable for grouped time series. Grouped time series are built based on a structure that disaggregates based on factors that can be both nested and crossed \cite{hyndman2018forecasting}. 

A group can have several levels, starting on the most aggregated level of the data and disaggregating it by attributes, e.g. A and B, forming the series $y_{\textrm{A},t}$ and $y_{\textrm{B},t}$, or by a second structure of attributes, e.g. X and Y, forming series $y_{\textrm{X},t}$ and $y_{\textrm{Y},t}$. At the bottom level, this would generate eight different series, $y_{\textrm{AX},t}$, $y_{\textrm{AY},t}$, $y_{\textrm{BX},t}$, $y_{\textrm{BY},t}$, $y_{\textrm{XA},t}$, $y_{\textrm{XB},t}$, $y_{\textrm{YA},t}$ and $y_{\textrm{YB},t}$.

\cite{hyndman2011optimal} proposed a new method to output more coherent forecasts, adding them up consistently within the aggregation structure. This method also proposes a generalized representation, from which the earlier approaches are obtained as special cases. As it is based on the reconciliation of the independent forecasts of all the aggregation levels, the reconciled forecasts are always more consistent than when a bottom-up strategy is used. This method is explored in more detail in the following section.

\subsection{Optimal Forecast Reconciliation}

The method consists in optimally combining and reconciling all forecasts at all levels of the hierarchy. To combine the independent forecasts, a linear regression is used to guarantee that the revised forecasts are as close as possible to the independent forecasts while maintaining coherency. The independent forecasts are reconciled  based on:
\begin{equation}
    \boldsymbol{\hat{y}}_{T}(h) = \boldsymbol{S} \, \boldsymbol{\beta}_{T}(h) + \epsilon_{h},
\end{equation} 
where $\boldsymbol{\hat{y}}_{T}(h)$ is a matrix of $h$-step-ahead independent forecasts for all series, stacked in the same order as the original data, $\boldsymbol{S}$ is a summing matrix, ${\boldsymbol{\beta}_{T} = \text{E}\,[\boldsymbol{b}_{T+h}\, |\, y_{1},...,y_{T}]}$ is the unknown mean of the most disaggregated series at the bottom level, and $\epsilon_{h}$ represents the reconciliation errors. \cite{hyndman2011optimal} also proved that the resulting forecasts are unbiased, which is not the case in the top-down approach.

\cite{hyndman2016fast} proposed a weighted least-squares (WLS) reconciliation approach, concluding that it outperforms the ordinary least squares (OLS). In their study, the only exception was the very top level where the OLS method did slightly better. The notation for the reconciliation approach was presented differently in \cite{hyndman2018optimal}, referring to the new method as the MinT (minimum trace) reconciliation, represented as:
\begin{equation}
    \boldsymbol{\Tilde{y}}_{T}(h) = \boldsymbol{S}\, (\boldsymbol{S}^{T} \, \boldsymbol{W}^{-1}_{h} \, \boldsymbol{S})^{-1} \, \boldsymbol{S}^{T}\, \boldsymbol{W}^{-1}_{h}\, \boldsymbol{\hat{y}}_{T}(h),
\end{equation}
where $\boldsymbol{\Tilde{y}}_{T}(h)$ is the reconciled $h$-step ahead forecast and $\boldsymbol{W}_{h}$ is the variance-covariance matrix of the $h$-step-ahead base forecast errors.
As it is challenging to estimate $\boldsymbol{W}_{h}$, \cite{hyndman2018optimal} proposed five different approximations for estimating this parameter, some of which we implemented in our work. 

Besides the MinT approach, we considered the approximation case where it is assumed that $W_{h} = k_{h}\it{I}, \forall\it{h}$, where $k_{h} > 0$ and \textit{I} is the identity matrix. In reality, this assumption collapses MinT to the early proposal of \cite{hyndman2011optimal}, to use a simple OLS estimator. This is a very strong assumption, where the base forecast errors are considered uncorrelated and equivariant, which is not possible to satisfy in hierarchical and grouped time series. 

We also considered the case where MinT can be described as a WLS estimator \cite{hyndman2018optimal}, which can be written as ${\boldsymbol{W_{h}}=k_{h}\text{diag}(\boldsymbol{\boldsymbol{\hat{W}}_{1}} )}$, for all $h$, where $k_{h}> 0$, and
    \begin{equation}
    \boldsymbol{\hat{W}_{1}}=\frac{1}{T}\sum_{t=1}^{T}{e_{t}e_{t}^{\prime}},
    \end{equation}
where $e_{t}$ is a vector of the residuals of the models that generated the base forecasts stacked in the same order as the data. This method scales the base forecasts using the variance of the residuals, hence it works as a weighted least squares estimator using variance scaling.

\section{PERFORMANCE MEASURES}\label{sec:performance_measures}
\medskip
\subsection{Mean Absolute Scaled Error (MASE)}

MASE is a measurement of forecast accuracy which scales the errors based on the \textit{na\"{i}ve} forecast method. If MASE is smaller than 1, the forecast method is better than the average one-step naive forecast. MASE is scale-free so can be used to compare different forecast accuracies. Its mathematical expression is given by:
\begin{equation}
    MASE = \text{mean}(|q|) = \frac{1}{T}\sum_{t=1}^{T}{|q|},
\end{equation}
where $q$ is defined as:
\begin{equation}
     q = \begin{cases}
         \ddfrac{ |\hat{y}_{t}-y_{t}|}{ \frac{1}{T-1}   \sum_{t=2}^{T}{| y_{t} - y_{t-1} |}     }\text{, if $\{y_{t}\}$ is non-seasonal}
         \\[30pt]
          \ddfrac{ |\hat{y}_{t}-y_{t}|}{ \frac{1}{T-s}   \sum_{t=s+1}^{T}{| y_{t} - y_{t-s} |}  } \text{, if $\{y_{t}\}$ is seasonal}, \\
        \end{cases}
\end{equation}
where $s$ is the seasonal period and $T$ is the length of the time series.

\subsection{Root Mean Squared Error (RMSE)}

RMSE is one of the most used metrics to evaluate forecast models. As it is not normalized, it can only be used to compare models in the same dataset. It is given by:
\begin{equation}
    RMSE = \sqrt{ \frac{1}{T} \sum_{t=1}^{T}{e_{t}^{2}}}. 
\end{equation}

\medskip
\section{DATASET}\label{sec:dataset}

HUUB is a logistics company for the fashion industry. It offers a platform for fashion brands that integrates the full supply chain process in one place, simplifying operations from production to customer service. HUUB's platform makes use of data to boost business growth and forecasting is a key aspect of its analysis. It has today over 50 brands under its monitoring, selling in 4 different channels:  eCommerce, marketplaces, stores and retail.

The time series of all HUUB brands have been analyzed in order to explore their sales behaviour. Only brands with more than two years of data and consistent sales on a weekly basis were considered for forecasting analysis. For an initial study, we selected only the eCommerce sales channel of two of the largest and most representative brands, ranked in the top 10. The aim of this data reduction is to narrow the scope of the problem, since the purpose is to have a first approach and a strong baseline for the future. Furthermore, combining only forecasts of the two most prominent brands helps to improve the forecast performance.  
\begin{figure}[h]
    \centering
    \includegraphics[trim={0 0.5cm 0 1.5cm},clip,width=\linewidth]{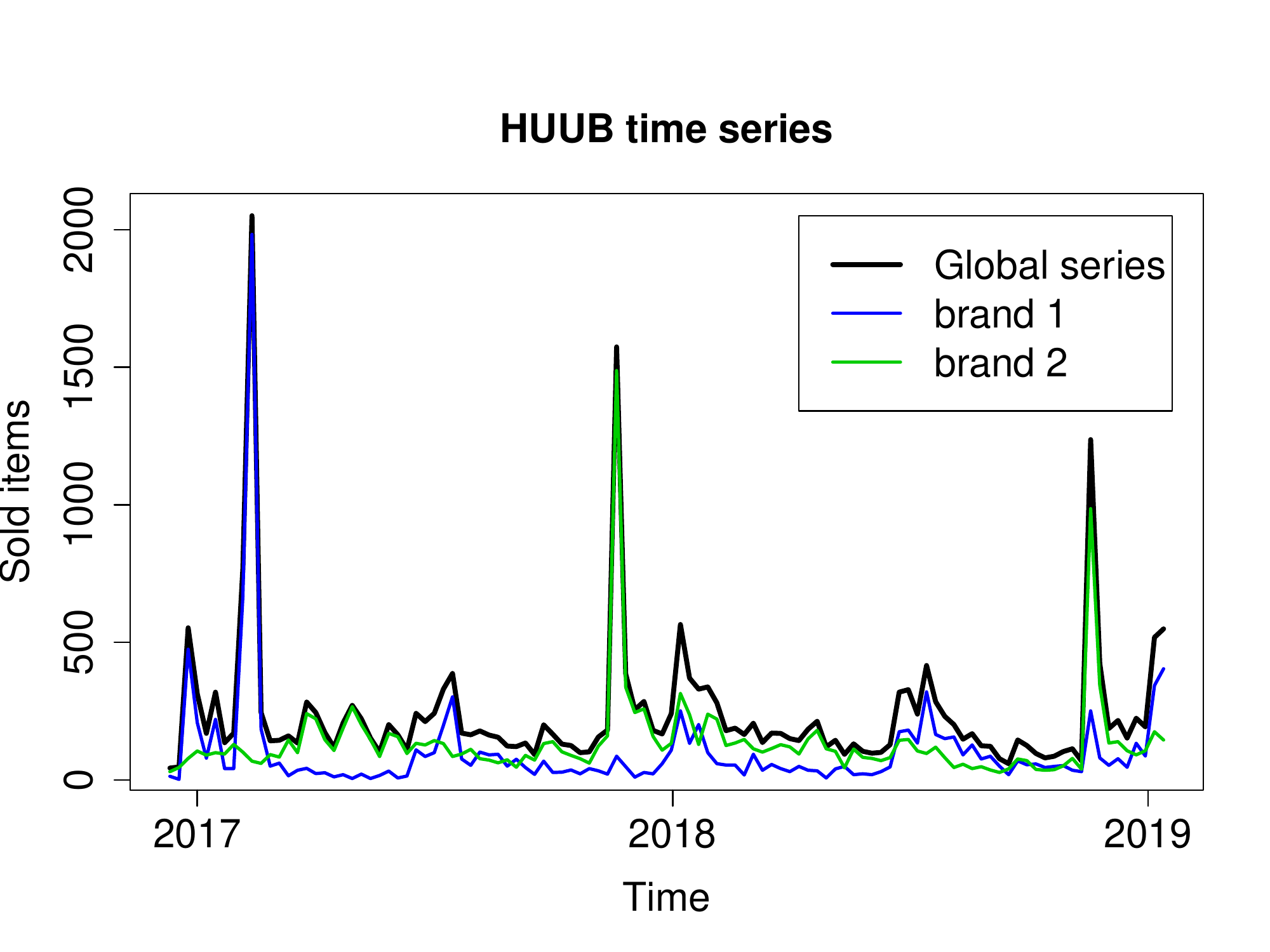}
    \caption{Time series profile of HUUB eCommerce selected brands.}
    \label{fig:brands_series}
\end{figure}

\setlength{\textfloatsep}{8pt}

Figure\,\ref{fig:brands_series} plots the selected eCommerce brands data series. The data is aggregated in weeks and ranges from the 11-12-2016 (Sunday), which is the earliest week for which we have complete data of all series included in the analysis, to the 19-01-2019 (Saturday).

\section{METHODS}\label{sec:methods}

\medskip
\subsection{Models specification}

\cite{box1976time} proposed a practical and pragmatic approach to build ARIMA models. \cite{Arunrajetal2016} defined, based on that methodology, 4 iterative steps, for the univariate time series SARIMA model.

\begin{enumerate}

    \item Model identification: This step involves the selection of parameters to include in the model, both seasonal and non-seasonal, based on the Autocorrelations function (ACF) and the Partial Autocorrelations function (PACF).

    \item Parameter estimation: the previously identified parameters are then estimated using Least Squares or Maximum Likelihood and a first selection of models is conducted using information criteria. 

    \item Diagnosis of the fitness of the model: The model is diagnosed using the statistical properties of the error terms, using Ljung-Box statistic test to check the adequacy. When the errors are normally distributed, we can move towards the next step.
    
    \item Forecasting and validation: The model is validated using out-of-sample data and applied to forecast the future values.

\end{enumerate}

\subsection{Grouped Time Series}

The structure used for this grouped time series starts at the top level with the time series that sums up all series at lower levels with different attributes. It is disaggregated by brand, \textit{brand\,1} and \textit{brand\,2}, in a hierarchical structure. Besides this, we have five different representations of the aggregation, since there are five attributes to consider in this study: gender (Male, Female or Unisex), age group (Baby, Kid, Baby Kid, Teen or Adult), product type (Clothing, Footwear, Accessories, Homewear, Beachwear, Swimwear, Swim accessories, Compound, Stationery, Underwear, Nightwear or Sports), season of the product (Seasonal or Permanent). 
\begin{figure}[h]
    \centering
    \includegraphics[width=0.9\linewidth]{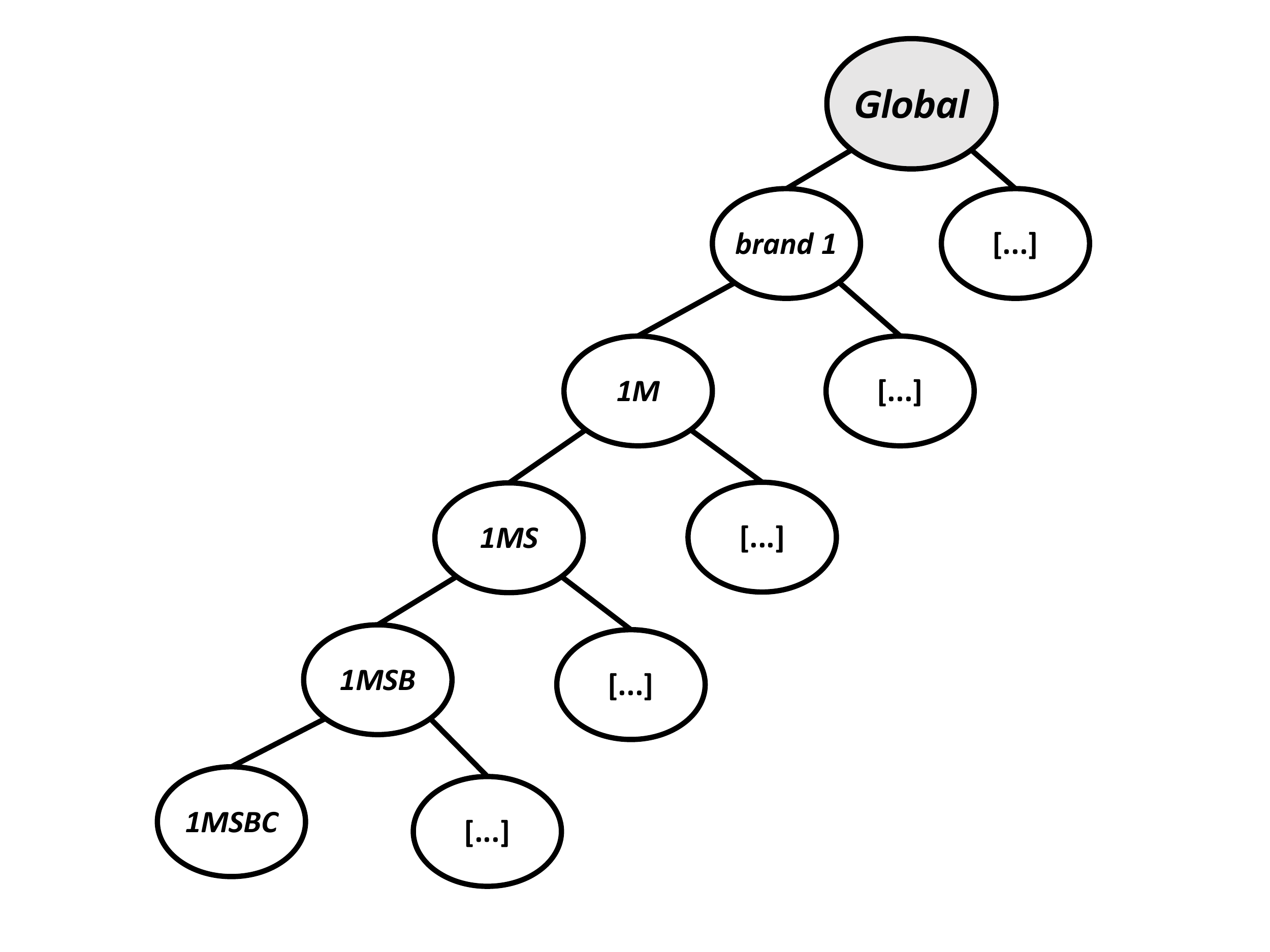}
    \caption{A representation of one of the five level grouped structures: (Global), the \textit{global} time series; \textit{brand\,1} (brand 1); \textit{brand\,1} crossed with gender \textit{male} (1M); \textit{brand\,1} crossed with gender \textit{male} and \textit{seasonal} items (1MS); \textit{brand\,1} crossed with gender \textit{male}, \textit{seasonal} items and age group \textit{baby} (IMSB); and, finally, \textit{brand\,1} crossed with gender \textit{male}, \textit{seasonal} items, age group \textit{baby} and product type \textit{clothing} (IMSBC).} \label{fig:grouped_timeseries}
\end{figure}
Figure~\ref{fig:grouped_timeseries} shows one of the alternatives to represent the group structure. It generates a total of 128 time series to be reconciled. When grouping time series, the complexity increases quickly, which requires significantly more computation resources to solve the problem. For each of the time series, we applied the best possible model to forecast that individual series using the \textit{auto.arima()} function from the R \textit{Forecast} package \cite{Hyndman2008,Hyndman2018}. 
\section{RESULTS}\label{sec:Results}

\subsection{Baseline model - SARIMA}

Based on the mathematical expression (\ref{eq:sarima}) for the SARIMA function, we can deduce some parameters restrictions according to the size of the time series window. For instance, if we have a time series of length 114 data points, and we choose a SARIMA model with parameters $(0,0,0)(2,1,0)^{52}$, this is an unfeasible model since it would be a linear combination of past data points given by:
\begin{equation}
    y_{t} = (1+\Phi_{1})\, y_{t-52} + (\Phi_{2}-\Phi_{1})\, y_{t-104} - \Phi_{2}\, y_{t-156}+ \epsilon_{t},
\end{equation}
and we do not have data for time period $t-156$, as would be required to compute the model.  

By working out the mathematical expression of the SARIMA model (Eq.~\ref{eq:sarima}) we can find the restrictions to the orders of the SARIMA model parameters as a function of the insample window size which is used to produce the forecasts. A forecast produced by a SARIMA model of parameters $(p,d,q)(P,D,Q)^{s}$ will only be feasible if the following conditions are met:
\begin{equation}\label{eq:restr}
        \boxed{
            \begin{aligned}
            &(D + P)\,s + p + d   \leq \text{window size}\\
            &Q\,s + q    \leq \text{window size}.
            \end{aligned}
              }
\end{equation}
Moreover, the best coefficients for each SARIMA model are found by fitting the model to the time series and in order to have all the data points to compute them, the time series should have at least one full seasonal period for which the past data points required to compute the series are all available. As an example, if we want to attempt fitting a SARIMA$(1,1,0)(0,1,0)^{52}$ model to a time series of length $T$, we will need at least $(D+P)s + p + d=54$ data points on top of a full seasonal period in order to have all the data points required to compute the model in at least one full seasonal period. If this is not the case, there will be some missing data and the model coefficients will not be as well-fit to the data as possible. The criteria used to estimate the quality of the model, such as the AIC and BIC, will not be well estimated in that case as well, given the missing data. 

Therefore, the more general restrictions for the SARIMA model as a function of the time series length are:
\begin{equation}\label{eq:restrictions}
        \boxed{
            \begin{aligned}
            &(D + P + 1)\,s + p + d   \leq T \\
            &(Q + 1)\,s + q   \leq T 
            \end{aligned}
              }
\end{equation}

Our time series has a length of $T=110$ data points (weeks) and a seasonal period of 52 weeks. We want to test a forecasting of 4 weeks, so we divide the time series into 106 data points for finding the best SARIMA parameters and coefficients, the training series, and 4 data points to test the forecast retrieved by that model, the testing series. The length of the training series limits our SARIMA parameters in the following way: ${(D + P)\,s + p + d \leq 54 }$ and ${Q\,s + q  \leq 54}$.

\medskip
\noindent{\textbf{\textit{Brand\,1}}}


The sales of \textit{brand\,1} vary a lot with time, which reflects in a time series with a high variance (see Figure~\ref{fig:brands_series}). This suggests that a Box-Cox transformation can make it more stationary. 

\begin{figure}[h]
    \centering
    \includegraphics[width=\linewidth]{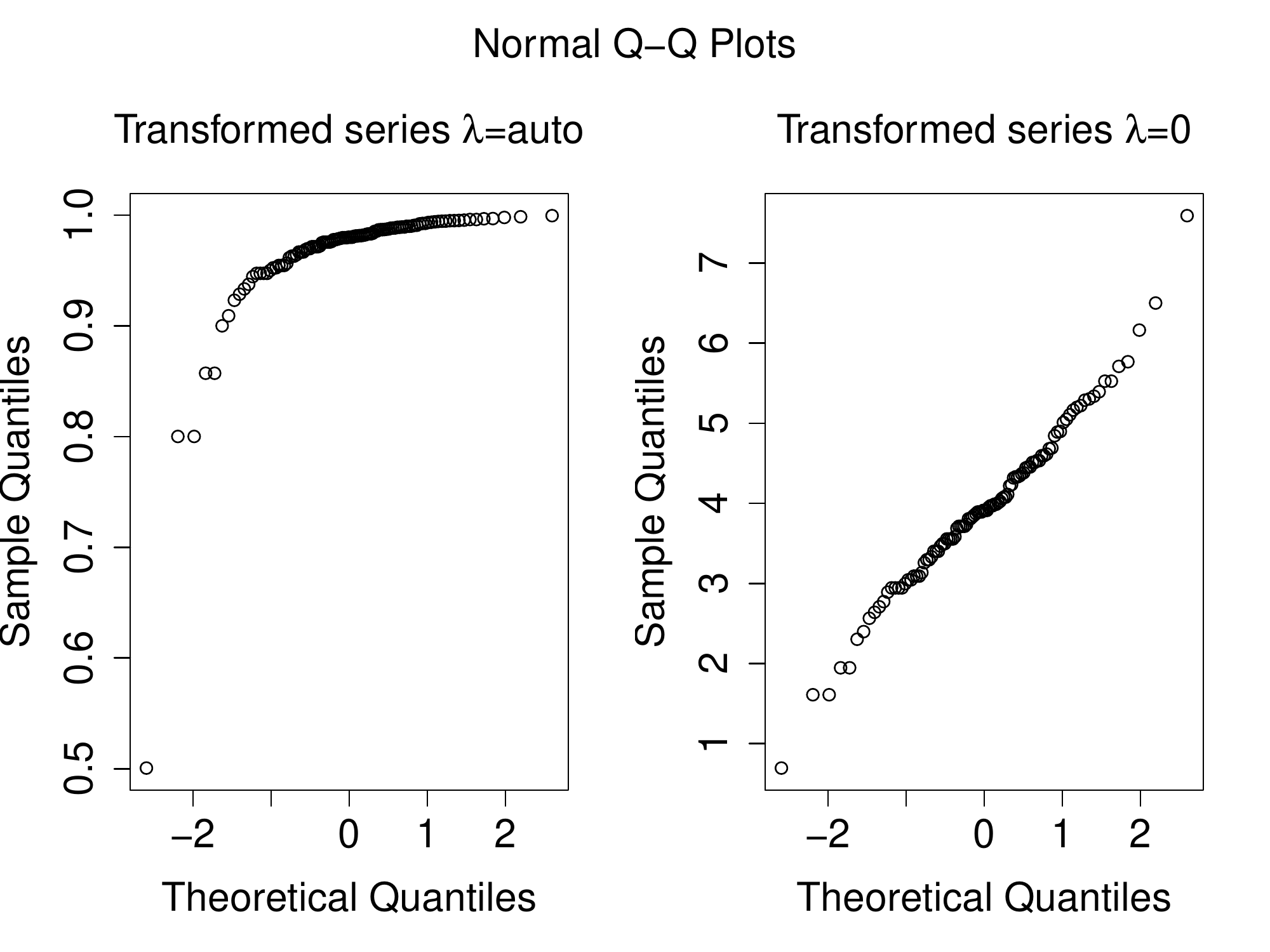}
    \caption{(Left:) Q-Q plot of \textit{brand\,1} Box-Cox transformed time series with $\lambda="auto"$; (Right:) Q-Q plot of \textit{brand\,1} Box-Cox transformed time series with $\lambda=0$, equivalent to a logarithmic transformation. It is clear that the latter series is closer to a normal distribution, hence it is a better transformation.} \label{fig:brand1_qq}
\end{figure}

The optimal $\lambda$ parameter can, in principle, be determined by ensuring that the standard deviation of the transformed data is minimum, which is what the \textit{BoxCox()} function from the R \textit{Forecast} package (\cite{Hyndman2008,Hyndman2018}) with \textit{lambda="auto"} computes. Even though there is a high likelihood that the data will be normally distributed after a Box-Cox transformation with \textit{lambda="auto"}, this is not guaranteed. A quantile-quantile (Q-Q) plot shows the distribution of the data against the expected normal distribution, hence it is an effective diagnosis to determine whether a sample is drawn from a normal distribution. Figure~\ref{fig:brand1_qq} shows the Q-Q plot of the Box-Cox transformed data with \textit{lambda="auto"} as well as the Q-Q plot of a simple logarithmic function, $\lambda=0$. The more data points fall in a straight line, the more normally distributed they are. If they deviate from a straight line in any systematic way, this suggests that the data is not drawn from a normal distribution. Figure~\ref{fig:brand1_qq} thus indicates that the logarithmic transformation effectively approximates the data to a normal distribution more than the Box-Cox transformation with \textit{lambda="auto"}. A ${\lambda=0}$ Box-Cox transformation was therefore applied to the data.

\begin{figure}[h]
    \centering
    \includegraphics[trim={0 0.5cm 0 2cm},clip,width=\linewidth]{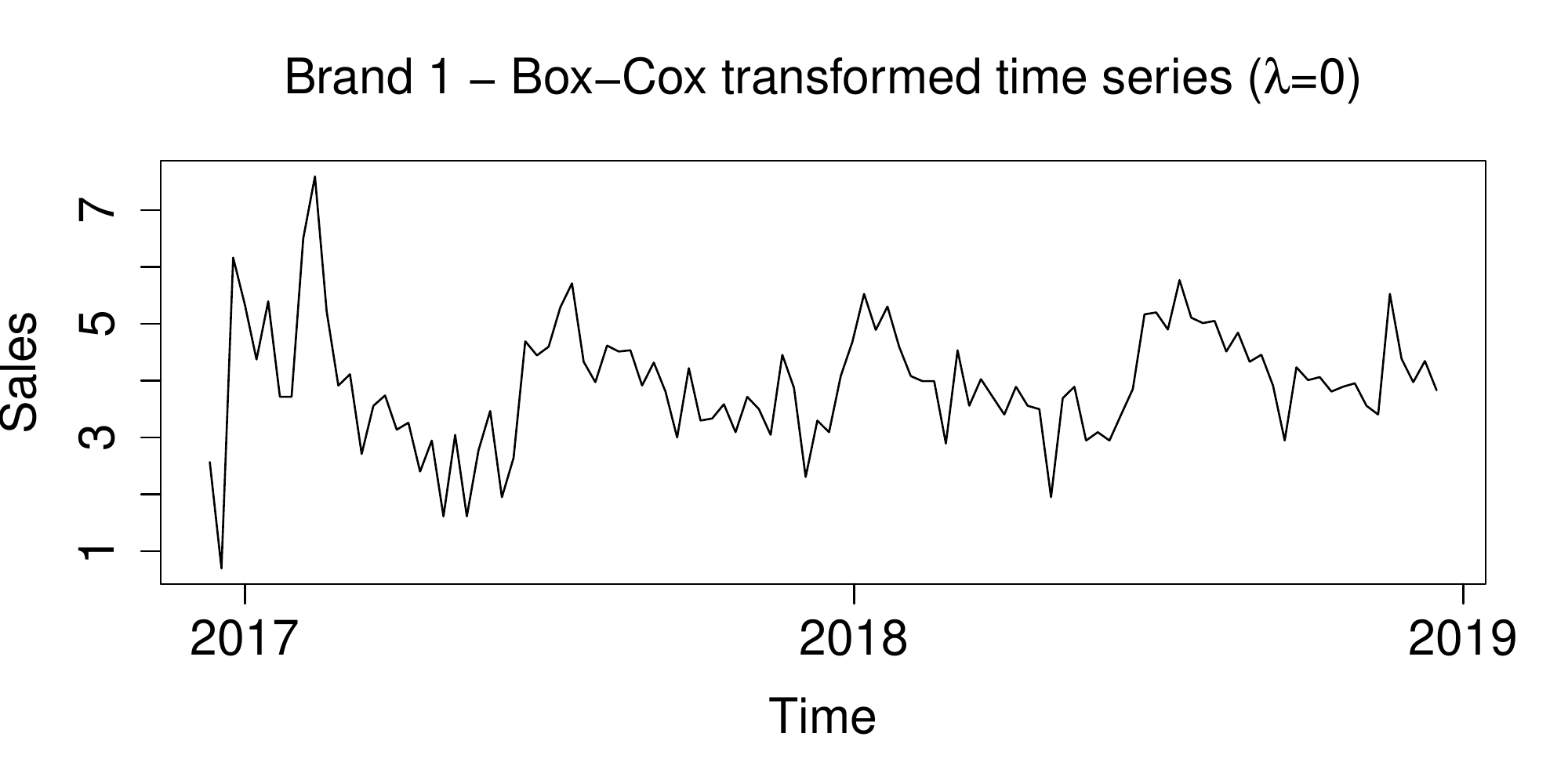}
    \caption{\textit{Brand\,1} time series after Box Cox transformation with $\lambda = 0$.} \label{fig:brand1_boxcox}
\end{figure}

\setlength{\textfloatsep}{8pt}

Figure~\ref{fig:brand1_boxcox} presents \textit{brand\,1}~1 time series after Box Cox transformation. This series displays a clear seasonal pattern with a cyclic profile with a yearly, or half-yearly, period. A seasonal differencing with a lag of 52 weeks (1 year) or 26 weeks (half-year) should, thus, improve the time series. Indeed, the time series standard deviation decreases from $1.053$ to $1.020$ and to $1.028$, after applying a 52-weeks and a 26-weeks differencing, respectively, which improves the series in both cases. A first differencing of the series increases the standard deviation and a unit root test also returns no first differencing, it is thus not expected to improve the modelling of the series. 

The analysis of the ACF and PACF of the yearly differenced series indicate that a SARIMA$(0,0,1)(0,1,0)^{52}$ model is a good candidate to reduce the data autocorrelations. This and other SARIMA models were tested and this was the model with the lowest AIC$_{\rm C}$ and BIC criteria. A more complex model, SARIMA$(6,0,0)(0,1,0)^{52}$, was favoured by the AIC criteria, however, according to the restrictions in (\ref{eq:restrictions}), this model exceeds the threshold limited by the size of the training series. Moreover, particularly when the sample size is small, the AIC criteria favours higher order (more complex) models as a result of overfitting the data. The AIC$_{\rm C}$ is a correction of the AIC and a more realistic criteria for smaller sample sizes, along with the BIC criteria. For these reasons, and since simpler models tend to be a better choice for accurately forecasting the general behaviour of time series, we chose $(0,0,1)(0,1,0)^{52}$. 

The half-year seasonally differenced series was also analysed and we found SARIMA$(0,0,0)(0,1,1)^{26}$ to be the best fit  model. Its RMSE was lower than most SARIMA models with a 52-weeks period, which could indicate that its forecast was better, but the AIC, AIC$_{\rm C}$ and BIC were significantly higher, implying that it provides a worst fit to the time series. The Ljung-Box test of the residuals of these models yielded lower p-values than for 52-weeks period models, signaling that they are worst models. For these reasons, we consider that the best SARIMA baseline model for \textit{brand\,1} is SARIMA$(0,0,1)(0,1,0)^{52}$.

\textit{Brand\,1} had a major peak of sales in the first quarter of 2017 which did not repeat in 2018 and which was almost $10$ times above the average of the subsequent peaks (see Figure\,\ref{fig:brands_series}). This sales peak was due to strong campaigning for an online collection launching. It was a one-time event and the peak affects the SARIMA models, increasing its prediction for the same period of subsequent years. We performed the same analysis having this outlier peak removed and its values interpolated. The time series forecasting RMSE was reduced by 22\%. However, to maintain the method's automation capacity, we chose not to implement any outlier removal. Instead, external variables will be used to handle this type of event in a subsequent article.

\medskip
\noindent{\textbf{\textit{Brand\,2}}}

\begin{figure}[h]
    \centering
    \includegraphics[width=\linewidth]{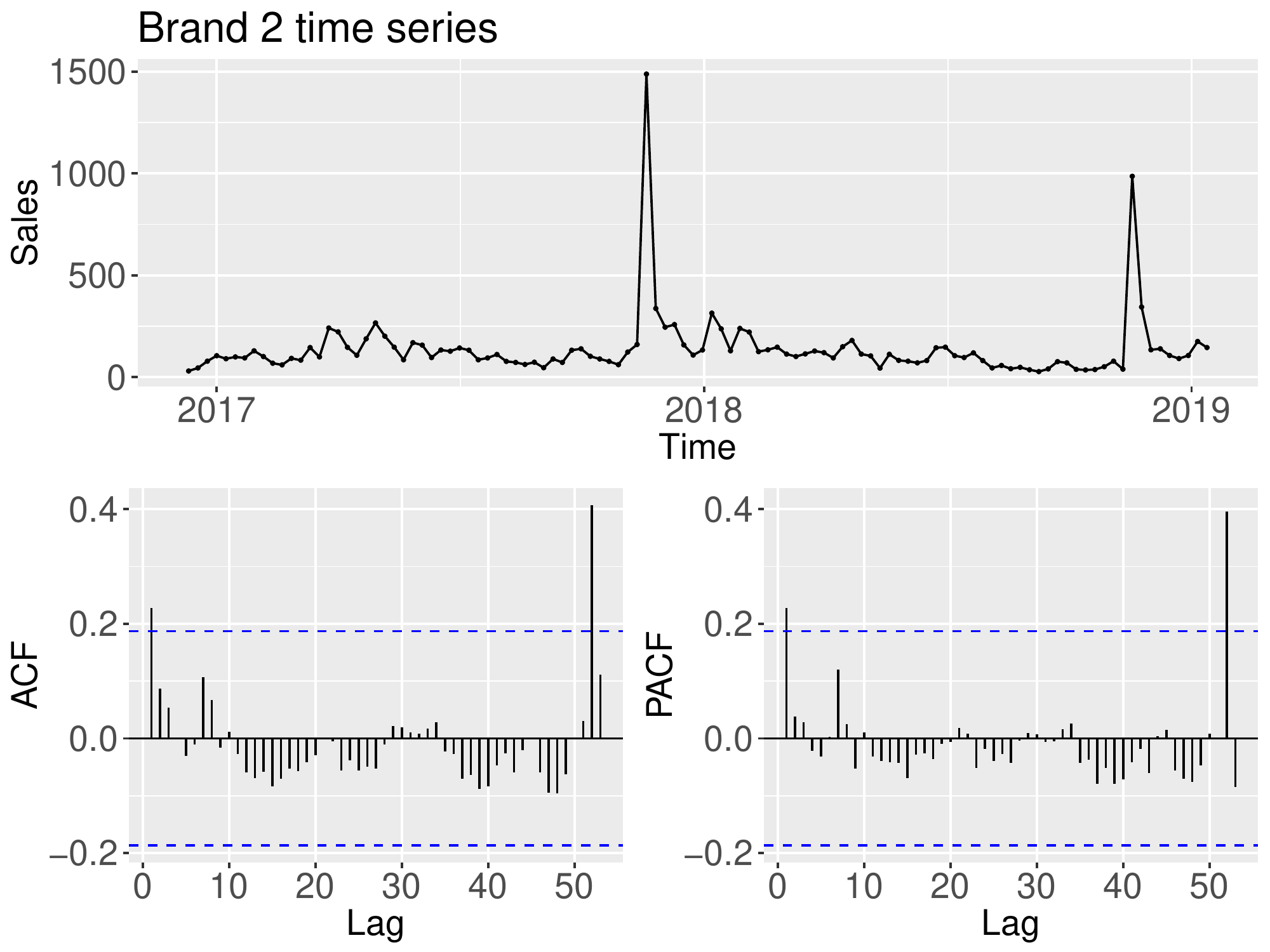}
    \caption{\textit{Brand\,2} original time series and respective ACF and PACF} \label{fig:brand2_original}
\end{figure}

The time series for \textit{brand\,2} has a seasonal profile, which is also clear in the ACF and PACF at lag\,52 (see Figure~\ref{fig:brand2_original}). Since the series shows variations which change with time, a transformation can improve the series. Similarly to \textit{brand\,1}, the Box-Cox transformation with $\lambda=0$ improved the normalization of the time series over $lambda="auto"$.

\setlength{\textfloatsep}{8pt}

Since \textit{brand\,2} displays a clear seasonal profile, it is not stationary and a seasonal difference can improve the series. Even though the standard deviation of the series increases from $0.64$ to $0.68$ after seasonal differencing, the series becomes more normalized. 

The series is also more stationary after a first difference and the standard deviation decreases to $0.54$. A second difference would increase significantly the standard deviation, so no more differencing was applied to the series. 


The ACF and PACF of the differenced series show a significant negative lag\,1 of $\approx -0.3$, which suggests a mild overdifferencing. This can be counterbalanced with one order of a moving average model applied to the series, which indicates that SARIMA$(0,1,1)(0,1,0)^{52}$ is a good candidate model for the data. 

This and other models, including different differencing orders, were tested and the BIC criteria, which favours simpler model, was used to compare them. The BIC criteria also favoured SARIMA$(0,1,1)(0,1,0)^{52}$ model out of all models. 


The residuals of the SARIMA$(0,1,1)(0,1,0)^{52}$ model applied to \textit{brand\,2} time series show all autocorrelation lags below the threshold, apart from lag 16, which corresponds to roughly a trimester and it is thus, most likely, related to the periodic behaviour of the brand. The fact that the residuals still display significant autocorrelations is also translated in a relatively low value of the LB p-value$=0.12$. As more data becomes available and are included in the time series, it will become more obvious whether the data autocorrelation at lag\,16 is real, and, if so, it will be easier to model it. 

\bigskip
\noindent{\textbf{Global time series}}

A global time series is formed by the sum of \textit{brand\,1} and \textit{brand\,2} time series. The same pre-processing and analysis performed on \textit{brand\,1} and \textit{2} time series were applied to the global time series, in the same 4 weeks period range.


As expected, the global time series also exhibits a seasonal profile, with a cyclic behaviour and significant autocorrelations at lags\,52. This entails a seasonal differencing, hence $D=1$. The standard deviation of the time series increases when applying a first difference and the unit root test determined that the time series is stationary so needs no first differencing, both advocating for $d=0$.
The model that presented lower AIC, AIC$_{c}$ and BIC criteria was SARIMA$(0,0,1)(0,1,0)^{52}$.


The residuals of SARIMA$(0,0,1)(0,1,0)^{52}$ showed no significant autocorrelations at any lag, which indicates that SARIMA$(0,0,1)(0,1,0)^{52}$ is a well-fit model for the global time series. Its residuals and the high LB p-value of $0.67$ indicate a reasonable resemblance of residuals with white noise, further approving this model.

\begin{figure}[h]
    \centering
    \includegraphics[width=\linewidth]{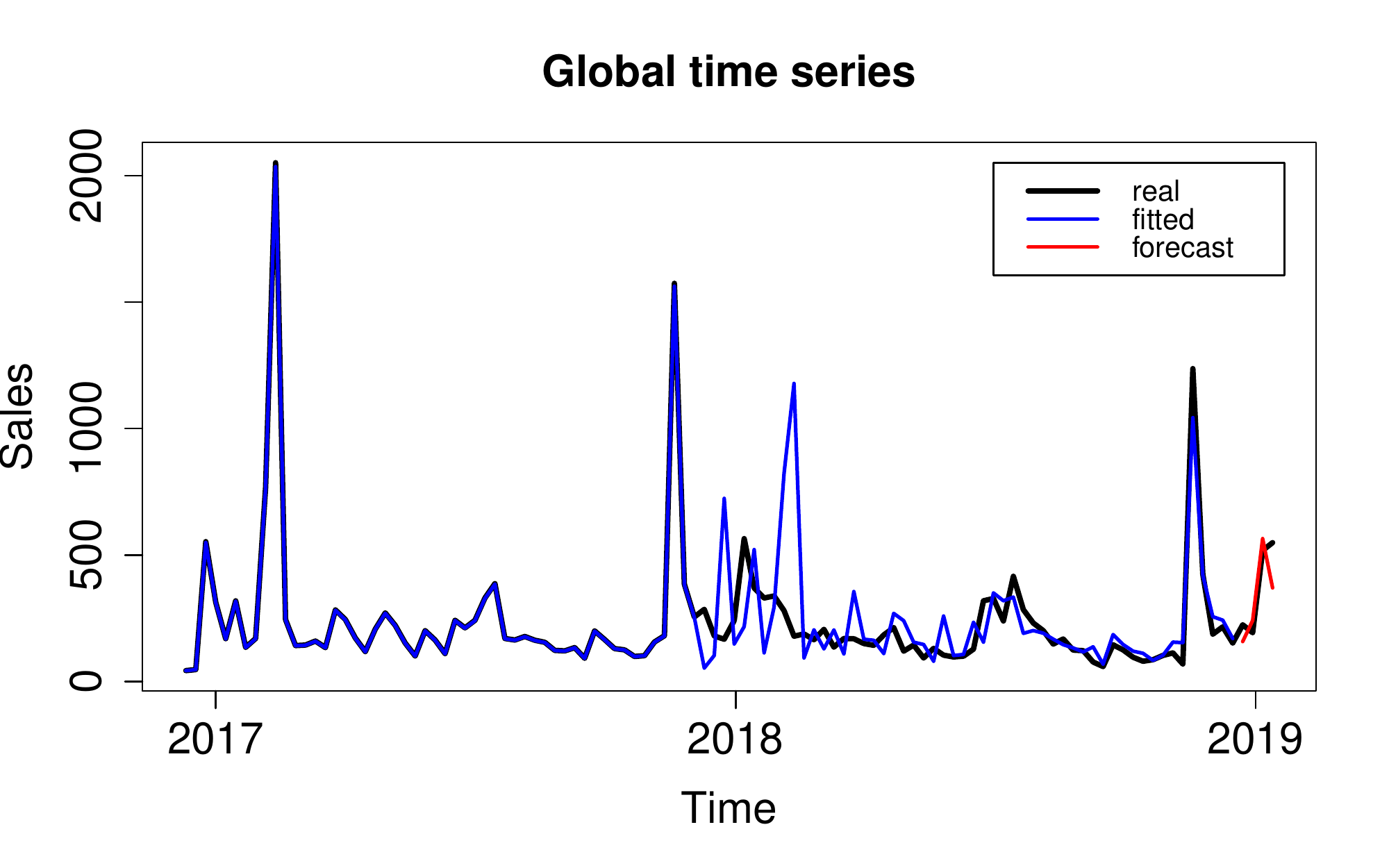}
    \caption{Global time series (black) with SARIMA$(0,1,1)(0,1,0)^{52}$  model fitting (blue) and 4-weeks forecasting (red).} \label{fig:global_result}
\end{figure}

The fitting and forecast of the global time series by SARIMA$(0,0,1)(0,1,0)^{52}$ are shown in Figure~\ref{fig:global_result}.

\medskip
\noindent{\textbf{Baseline models performance}}

The baseline models performance summary is presented in Table~\ref{tab:global_forecast_models} 

\begin{table}[h]
    \caption{Baseline SARIMA models for \textit{brand\,1}, \textit{brand\,2} and the \textit{global} time series. The time period of forecast was 4 weeks. }\label{tab:global_forecast_models}
\begin{center}
\begin{tabular}{|@{\hspace{0.3\tabcolsep}}c@{\hspace{0.3\tabcolsep}}|@{\hspace{0.3\tabcolsep}}c@{\hspace{0.3\tabcolsep}}|@{\hspace{0.3\tabcolsep}}c@{\hspace{0.3\tabcolsep}}|@{\hspace{0.3\tabcolsep}}c@{\hspace{0.3\tabcolsep}}|@{\hspace{0.3\tabcolsep}}c@{\hspace{0.3\tabcolsep}}|@{\hspace{0.3\tabcolsep}}c@{\hspace{0.3\tabcolsep}}|@{\hspace{0.3\tabcolsep}}c@{\hspace{0.3\tabcolsep}}|@{\hspace{0.3\tabcolsep}}c@{\hspace{0.3\tabcolsep}}|}
\hline
    \textbf{Data} & \textbf{SARIMA} & \textbf{AIC} & \textbf{AIC$_{c}$} & \textbf{BIC} & \textbf{LB} & \textbf{MASE} & \textbf{RMSE} \\
     & & & & & \textbf{p-value} & & \\
\hline
    Global & $(0,0,1)(0,1,0)^{52}$ & 101.05 & 101.28 & 105.02 & 0.67 & 0.69 & 101.16  \\
    \textit{Brand\,1} & $(0,0,1)(0,1,0)^{52}$  & 153.72 & 153.95 & 157.70 & 0.80 & 1.17 & 145.79\\
    \textit{Brand\,2} & $(0,1,1)(0,1,0)^{52}$ &  77.23 & 77.47 & 81.17 & 0.12 &  0.24 & 19.48\\
\hline
\end{tabular}
\end{center}
\end{table}


\subsection{Grouped Time Series Forecast Reconciliation}

As explained in Section~\ref{sec:methods}, the five different product attributes selected generated 128 different time series to reconciliate. The best possible model parameters and coefficients were determined for each of these individual series, using the \textit{auto.rima()} function \cite{Hyndman2008,Hyndman2018}. The best individual forecasts were used as input for the reconciliation process. Table~\ref{tab:global_perform_group} shows the performance metrics in the out-of-sample data: independent (base) forecasts, used as baseline, bottom-up and the three approaches considered as the optimal reconciliation methods (OLS, WLS and MinT, proposed by \cite{hyndman2016fast} and \cite{hyndman2018optimal}). 

All approaches were tested using a horizon $h$ of 4 weeks, which was identified by HUUB as the most relevant period range to be forecasted, given the weekly data granularity. The results are presented for the three most aggregated time series, despite the fact that the grouped time series yielded results for all the possible combinations. This is also a major advantage for the business, allowing to drill down to more granular dimensions, with the reassurance that the forecasts are consistent across these dimensions. The decision making process can be significantly empowered, for instance when defining purchase quantities for the next collection for specific product type, age group, etc.

Overall, the results show that the best performing method for each series can leverage nested information in the group structure and add consistency to the outcome. This is supported by the RMSE of WLS in \textit{brand\,1}, which has a very significant improvement when compared to the baseline. In terms of \textit{brand\,2}, WLS results were very close to the baseline, which can be explained by the simpler grouping structure for this brand, i.e. it is essentially concentrated in one product type, one gender, one age group, one season. For the \textit{global} time series, the baseline model performed slightly better than the reconciled best model, with an RMSE 5\% lower than WLS. Overall, the RMSE indicates that WLS is a robust method for the forecasting of seasonal time series when no further information is known about what causes the time series variations.

The best performing model in terms of MASE metric is the simpler OLS method for \textit{brand\,1} and \textit{global} time series. For \textit{brand\,2}, MASE is lower for the baseline model. The MASE error of the WLS is lower than that of the baseline for the \textit{global} and \textit{brand\,1} series, and very similar for \textit{brand\,2}. 

Adding more brands and more attributes can, in principle, help to increase the accuracy of the proposed models relative to the baseline and, indeed, preliminary work with a disaggregation with more attributes is showing a significant improvement. 

Examining the proposed methods, the one that yielded better results in terms of RMSE was the WLS, outperforming the very ineffective bottom-up, but also the simpler OLS and the new MinT approach. 

\begin{table}[h]
    \caption{Models performance in the out-of-sample data of the baseline forecasts, bottom-up and the approaches considered as the optimal reconciliation methods: OLS, WLS and MinT. The best performing models according to the RMSE are highlighted in bold. }\label{tab:global_perform_group}
\begin{center}
\begin{tabular}{|c||c||c||c|}
\hline
    \textbf{Data} & \textbf{Model} & \textbf{MASE} & \textbf{RMSE}  \\
\hline
\hline
\multirow{5}{*}{\textit{Global}}    &\textbf{Baseline} & \textbf{0.69} & \textbf{101.16}\\
    &Bottom-up & 1.16 & 180.93  \\
    &OLS & 0.63 & 108.08 \\
    &WLS & 0.67 & 105.96 \\
    &MinT & 0.72 & 109.51 \\
\hline
\hline
\multirow{5}{*}{\textit{Brand\,1}}    &Baseline & 1.17 & 145.79  \\
    &Bottom-up & 1.74 & 189.41  \\
    &OLS & 0.81 & 113.21 \\
    &\textbf{WLS} & \textbf{0.86} & \textbf{107.72} \\
    &MinT & 0.94 & 110.21 \\
\hline
\hline
\multirow{5}{*}{\textit{Brand\,2}}    &\textbf{Baseline} &  \textbf{0.24} & \textbf{19.48} \\
    &Bottom-up &  0.31 & 26.07  \\
    &OLS &  0.30 & 22.33 \\
    &\textbf{WLS} &  \textbf{0.26} & \textbf{19.55} \\
    &MinT &  0.25 & 20.18 \\
\hline
\end{tabular}
\end{center}
\end{table}


\vspace{-0.5cm}
\section{CONCLUSIONS AND FUTURE WORK}\label{sec:conclusion}

This work explored the optimal forecasts method for hierarchical and grouped time series proposed by \cite{hyndman2011optimal,hyndman2018optimal} to forecast the weekly sales of two fashion brands under HUUB's monitoring. The data were disaggregated according to five selected attributes: brand, gender, whether the item is seasonal or permanent, age group and product type. This disaggregation resulted in a grouped structure with 128 individual time series. These series were individually fit by the best SARIMA model found by the \textit{auto.arima()} function and were then reconciled following the optimal forecasts method. Three difference cases of the optimal forecasts method were explored: OLS, WLS and MinT, as well as the simple Bottom-up approach.

The forecasts performance of the three most aggregate series was evaluated and compared to the forecasts obtained by baseline SARIMA models. Each of these baseline SARIMA model was found by manually analysing the time series ACF, PACF and through stationarity and differencing tests in order to determine the best model parameters and coefficients. 

The comparison of the performance of the optimal forecasts method and the baseline SARIMA models showed that WLS was the forecast method with the lowest RMSE for the two brands time series. The global time series was slightly better forecasted by the baseline model. Overall, the RMSE indicates that WLS is a robust method for the forescasting of seasonal time series when no further information is known about what causes the time series variations. Preliminary work with a higher-level disaggregation with more attributes has shown a significant improvement of the RMSE of the global time series forecast. These findings will be presented in a subsequent article.

Out of the three explored methods, WLS was the one with the highest RMSE accuracy followed by MinT and then OLS. All of them performed better than the less efficient bottom-up method, as expected.

\medskip
The proposed work was intentionally focused on univariate time series models, providing a strong baseline and framework to build more complex forecasting models. Future work will contemplate the addition of more brands and more attributes (not only product attributes but others like sales markets geography structures), increasing the capacity of the model to capture even more nested information in these granular series. 
One possibility is the use of hybrid models, that combine statistics and machine learning, already proven to generate better results than statistical ones \cite{Makridakisetal2018}. 

Another aspect that our work will exploit is the use of multivariate models, since they can add significant information on the volatile and unexpected behaviour observed in the data.

\addtolength{\textheight}{-12cm}   









\bibliographystyle{IEEEtran}
\bibliography{IEEEabrv,main}

\begin{thebibliography}{10}
\providecommand{\url}[1]{#1}
\csname url@samestyle\endcsname
\providecommand{\newblock}{\relax}
\providecommand{\bibinfo}[2]{#2}
\providecommand{\BIBentrySTDinterwordspacing}{\spaceskip=0pt\relax}
\providecommand{\BIBentryALTinterwordstretchfactor}{4}
\providecommand{\BIBentryALTinterwordspacing}{\spaceskip=\fontdimen2\font plus
\BIBentryALTinterwordstretchfactor\fontdimen3\font minus
  \fontdimen4\font\relax}
\providecommand{\BIBforeignlanguage}[2]{{%
\expandafter\ifx\csname l@#1\endcsname\relax
\typeout{** WARNING: IEEEtran.bst: No hyphenation pattern has been}%
\typeout{** loaded for the language `#1'. Using the pattern for}%
\typeout{** the default language instead.}%
\else
\language=\csname l@#1\endcsname
\fi
#2}}
\providecommand{\BIBdecl}{\relax}
\BIBdecl

\bibitem{Lenort2013}
R.~Lenort and P.~Besta, ``{Hierarchical sales forecasting system for apparel
  companies and supply chains},'' \emph{Fibres and Textiles in Eastern Europe},
  vol.~21, no.~6, pp. 7--11, 2013.

\bibitem{Capistran2010}
C.~Capistr{\'{a}}n, C.~Constandse, and M.~Ramos-Francia, ``{Multi-horizon
  inflation forecasts using disaggregated data},'' \emph{Economic Modelling},
  vol.~27, no.~3, pp. 666--677, 2010.

\bibitem{athanasopoulos2009hierarchical}
G.~Athanasopoulos, R.~A. Ahmed, and R.~J. Hyndman, ``Hierarchical forecasts for
  australian domestic tourism,'' \emph{International Journal of Forecasting},
  vol.~25, no.~1, pp. 146--166, 2009.

\bibitem{Shang2017}
H.~L. Shang and S.~Haberman, ``{Grouped multivariate and functional time series
  forecasting: An application to annuity pricing},'' \emph{Insurance:
  Mathematics and Economics}, vol.~75, pp. 166--179, 2017.

\bibitem{hyndman2011optimal}
R.~J. Hyndman, R.~A. Ahmed, G.~Athanasopoulos, and H.~L. Shang, ``Optimal
  combination forecasts for hierarchical time series,'' \emph{Computational
  Statistics \& Data Analysis}, vol.~55, no.~9, pp. 2579--2589, 2011.

\bibitem{hyndman2016fast}
R.~J. Hyndman, A.~J. Lee, and E.~Wang, ``Fast computation of reconciled
  forecasts for hierarchical and grouped time series,'' \emph{Computational
  Statistics \& Data Analysis}, vol.~97, pp. 16--32, 2016.

\bibitem{Makridakis2018}
S.~Makridakis, E.~Spiliotis, and V.~Assimakopoulos, ``Statistical and machine
  learning forecasting methods: Concerns and ways forward,'' \emph{PLoS ONE},
  vol.~13, 03 2018.

\bibitem{Makridakisetal2018}
------, ``{The M4 Competition: Results, findings, conclusion and way
  forward},'' \emph{International Journal of Forecasting}, vol.~34, no.~4, pp.
  802--808, 2018.

\bibitem{2013arXiv1302}
R.~Adhikari and R.~K. Agrawal, ``An introductory study on time series modeling
  and forecasting,'' \emph{CoRR}, vol. abs/1302.6613, 2013.

\bibitem{hyndman2018forecasting}
R.~J. Hyndman and G.~Athanasopoulos, \emph{Forecasting: principles and
  practice}.\hskip 1em plus 0.5em minus 0.4em\relax OTexts, 2018.

\bibitem{ljungbox1978}
G.~M. LJUNG and G.~E.~P. BOX, ``On a measure of lack of fit in time series
  models,'' \emph{Biometrika}, vol.~65, no.~2, pp. 297--303, 1978.

\bibitem{fliedner2001hierarchical}
G.~Fliedner, ``Hierarchical forecasting: issues and use guidelines,''
  \emph{Industrial Management \& Data Systems}, vol. 101, no.~1, pp. 5--12,
  2001.

\bibitem{hyndman2018optimal}
S.~L~Wickramasuriya, G.~Athanasopoulos, and R.~Hyndman, ``Optimal forecast
  reconciliation for hierarchical and grouped time series through trace
  minimization,'' \emph{Journal of the American Statistical Association}, pp.
  1--45, 03 2018.

\bibitem{box1976time}
G.~E. Box and G.~M. Jenkins, \emph{Time series analysis: forecasting and
  control, revised ed}.\hskip 1em plus 0.5em minus 0.4em\relax Holden-Day,
  1976.

\bibitem{Arunrajetal2016}
N.~S. Arunraj, D.~Ahrens, and M.~Fernandes, ``Application of {SARIMAX} model to
  forecast daily sales in food retail industry,'' \emph{{IJORIS}}, vol.~7,
  no.~2, pp. 1--21, 2016.

\bibitem{Hyndman2008}
R.~J. Hyndman and Y.~Khandakar, ``Automatic time series forecasting: the
  forecast package for {R},'' \emph{Journal of Statistical Software}, vol.~26,
  no.~3, pp. 1--22, 2008.

\bibitem{Hyndman2018}
R.~Hyndman, G.~Athanasopoulos, C.~Bergmeir, G.~Caceres, L.~Chhay,
  M.~O'Hara-Wild, F.~Petropoulos, S.~Razbash, E.~Wang, and F.~Yasmeen,
  \emph{{forecast}: Forecasting functions for time series and linear models},
  2018, r package version 8.4.

\end{thebibliography}

\end{document}